\documentclass[runningheads]{llncs}
\usepackage{graphicx}
\usepackage{amsmath}
\usepackage{booktabs}
\usepackage[colorlinks, linkcolor=blue, anchorcolor=blue, citecolor=blue]{hyperref}

\newcommand{\etal}{\emph{et al.}}
\newcommand{\ie}{\emph{i.e.},}
\newcommand{\eg}{\emph{e.g.},}

\begin{document}
\title{Liver Tumor Localization and Characterization from Multi-Phase MR Volumes Using Key-Slice Parsing: A Physician-Inspired Approach}
\titlerunning{Liver Tumor Localization and Characterization}
\author{Bolin Lai\inst{1}\thanks{B. Lai, Y. Wu and X. Bai --- Equal contribution. \\This work was done when X. Bai was intern at Ping An Technology.} \and Yuhsuan Wu\inst{1 \star} \and Xiaoyu Bai\inst{1,2 \star} \and Xiao-Yun Zhou\inst{3} \and Peng Wang\inst{5} \and Jinzheng Cai\inst{3} \and Yuankai Huo\inst{4} \and Lingyun Huang\inst{1} \and Yong Xia\inst{2} \and Jing Xiao\inst{1} \and Le Lu\inst{3} \and Heping Hu\inst{5} \and Adam P. Harrison\inst{3}.}
\authorrunning{B. Lai et al.}
%
\institute{Ping An Technology, Shanghai, China \and
Northwestern Polytechnical University, Xian China \and
PAII Inc., Bethesda, Maryland, USA \and
Vanderbilt University, Nashville, Tennessee, USA \and
Eastern Hepatobiliary Surgery Hospital, Shanghai, China}

\maketitle              
\begin{abstract}
Using radiological scans to identify liver tumors is crucial for proper patient treatment. This is highly challenging, as top radiologists only achieve F1 scores of roughly $80\%$ (hepatocellular carcinoma (HCC) vs. others) with only moderate inter-rater agreement, even when using multi-phase magnetic resonance (MR) imagery. Thus, there is great impetus for computer-aided diagnosis (CAD) solutions. A critical challenge is to \emph{robustly} parse a 3D MR volume to localize diagnosable regions of interest (ROI), especially for edge cases. In this paper, we break down this problem using a key-slice parser (KSP), which emulates physician workflows by first identifying key slices and then localizing their corresponding key ROIs. To achieve robustness, the KSP also uses curve-parsing and detection confidence re-weighting. We evaluate our approach on the largest multi-phase MR liver lesion test dataset to date ($430$ \emph{biopsy-confirmed} patients). Experiments demonstrate that our KSP can localize diagnosable ROIs with high reliability: $87\%$ patients have an average 3D overlap of $>=40\%$ with the ground truth compared to only $79\%$ using the best tested detector. When coupled with a classifier, we achieve an HCC vs. others F1 score of $0.801$, providing a fully-automated CAD performance  comparable to top human physicians.

\end{abstract}
\section{Introduction}
\noindent Liver cancer is the fifth/eighth most common malignancy in men/women worldwide~\cite{bosch2004primary}. During treatment planning, non-invasive diagnostic imaging is preferred, as invasive procedures, \ie{} biopsies or surgeries, can lead to hemmorages, infections, and even death~\cite{grant1999guidelines}. Multi-phase magnetic resonance (MR) imagery is  considered the most informative radiological option~\cite{oliva2004liver}, with T2-weighted imaging (T2WI) able to reveal tumor edges and aggressiveness~\cite{aube2017easl}. Manual lesion differentiation is workload-heavy and ideally is executed by highly experienced radiologists that are not always available in every medical center. Studies on human reader performance, which focus on differentiating hepatocellular carcinoma (HCC) from other types, report low specificities~\cite{aube2017easl} and moderate inter-rater agreement~\cite{fowler_inter_rater_variability}. Thus, there is a need for computer-aided diagnosis (CAD) solutions, which is the topic of our work. Unlike other approaches, we propose a physician-inspired workflow to achieve greater reliability and robustness. 

A major motivation for CAD is addressing challenging cases that would otherwise be biopsied or even incorrectly operated on. For instance, a 2006 retrospective study discovered that pre-operative imaging misinterpreted $20\%$ of its liver transplant patients as having HCC~\cite{freeman2006optimizing}. While several CAD approaches have been reported, many do not focus on histopathologically-confirmed studies~\cite{yang2012content,adcock2014classification,diamant2015improved,wu2019radiomics}, which are the cases most requiring CAD intervention. Prior CAD studies, except for Zhen \etal{}~\cite{zhen2020deep}, also only focus on computed tomography (CT), despite the greater promise of MR.   Most importantly, apart from two studies~\cite{chen2019cascade,huo2020harvesting}, CAD works typically assume a manually drawn region of interest (ROI) is available. In doing so, they elide the major challenge of parsing a medical volume to determine diagnosable ROIs. Without this capability, manual intervention remains necessary and the system also remains susceptible to inter-user variations. The most obvious localization strategy, \eg{} that of \cite{huo2020harvesting}, would follow computer vision practices and directly applies a detector. However, detectors aim to find \emph{all} lesions and their \emph{entire} 3D extent in a study, whereas the needs for liver lesion characterization are distinct: \emph{reliably} localize one or more key diagnosable ROI(s). This different goal warrants its own study, which we investigate.

\begin{figure}[t]
\centering
\includegraphics[width=0.9\columnwidth]{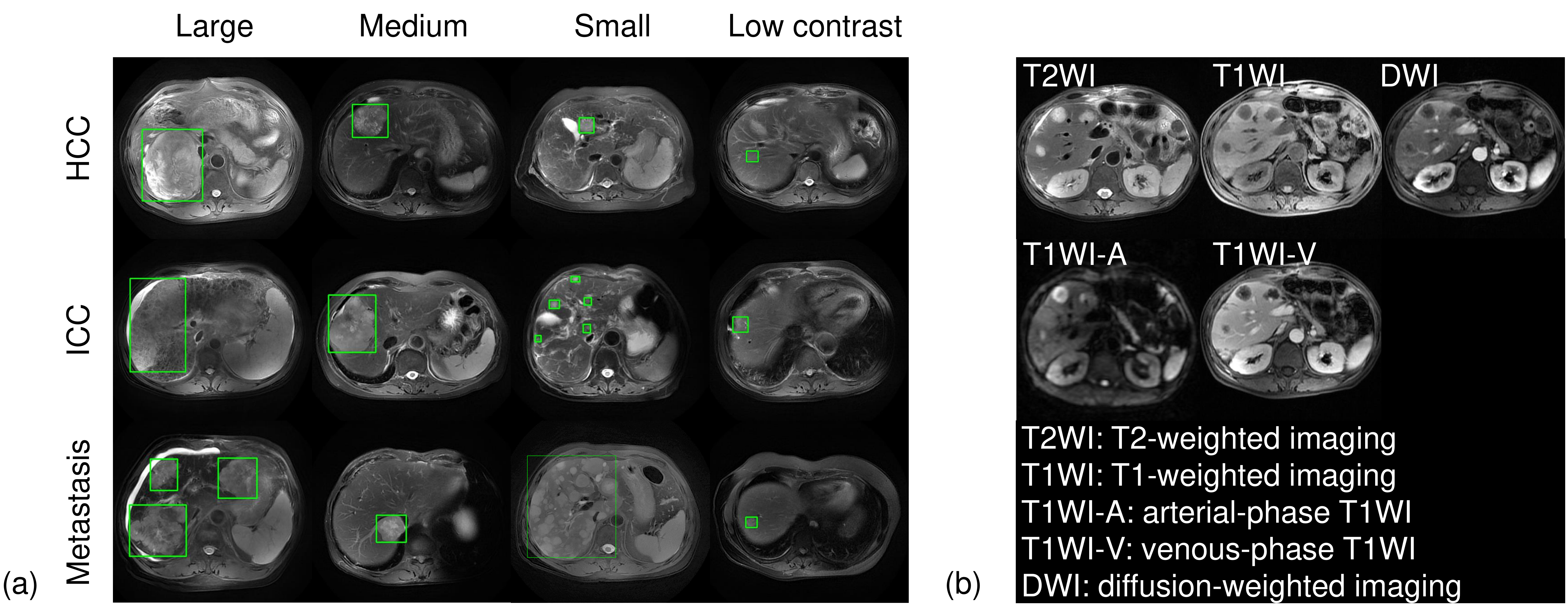}
\caption{(a) HCC, ICC and metastasis lesion examples on T2WIs, including large, medium, small and low contrast tumors. The ``small'' metastasis shows an example of a lesion cluster. (b) Different MR sequences of the same patient.}
\label{fig: tumor examples}
\end{figure}

In this work, we develop a robust and fully-automated CAD system to differentiate malignant liver tumors into the HCC, ICC and metastasis subtypes. Fig.~\ref{fig: tumor examples}(a) depicts these three types. To localize key diagnosable ROIs, we use a physician-inspired approach that departs from standard detection frameworks seen in computer vision and used elsewhere~\cite{huo2020harvesting}. Instead, we propose a key-slice parser (KSP), which breaks down the parsing problem similarly to clinical practice, \ie{} first robustly identifying and ranking key slices in the volume and, from each of these, regressing a single diagnosable ROI. This follows, at least in spirit, protocols like the ubiquitous response evaluation criteria in solid tumors (RECIST)~\cite{Eisenhauer2009}. In concrete terms KSP comprises multi-sequence classification, detection, and curve parsing. Once localized, each ROI is classified using a standard classifier.

We test our approach on $430$ multi-phase MR studies ($2150$ scans), \emph{which is the largest test cohort studied for liver lesion CAD to date.} Moreover, all of our patient studies are histopathologically confirmed, well-representing the challenging cases requiring CAD. Using our KSP framework, we achieve very high reliability, with $87\%$ of our predicted ROIs overlapping with the ground truth by $>=40\%$, outperforming the best detector alternative (only $79\%$ with an overlap $>=40\%$).  

\begin{figure}[t]
\centering
\includegraphics[width=1.0\columnwidth]{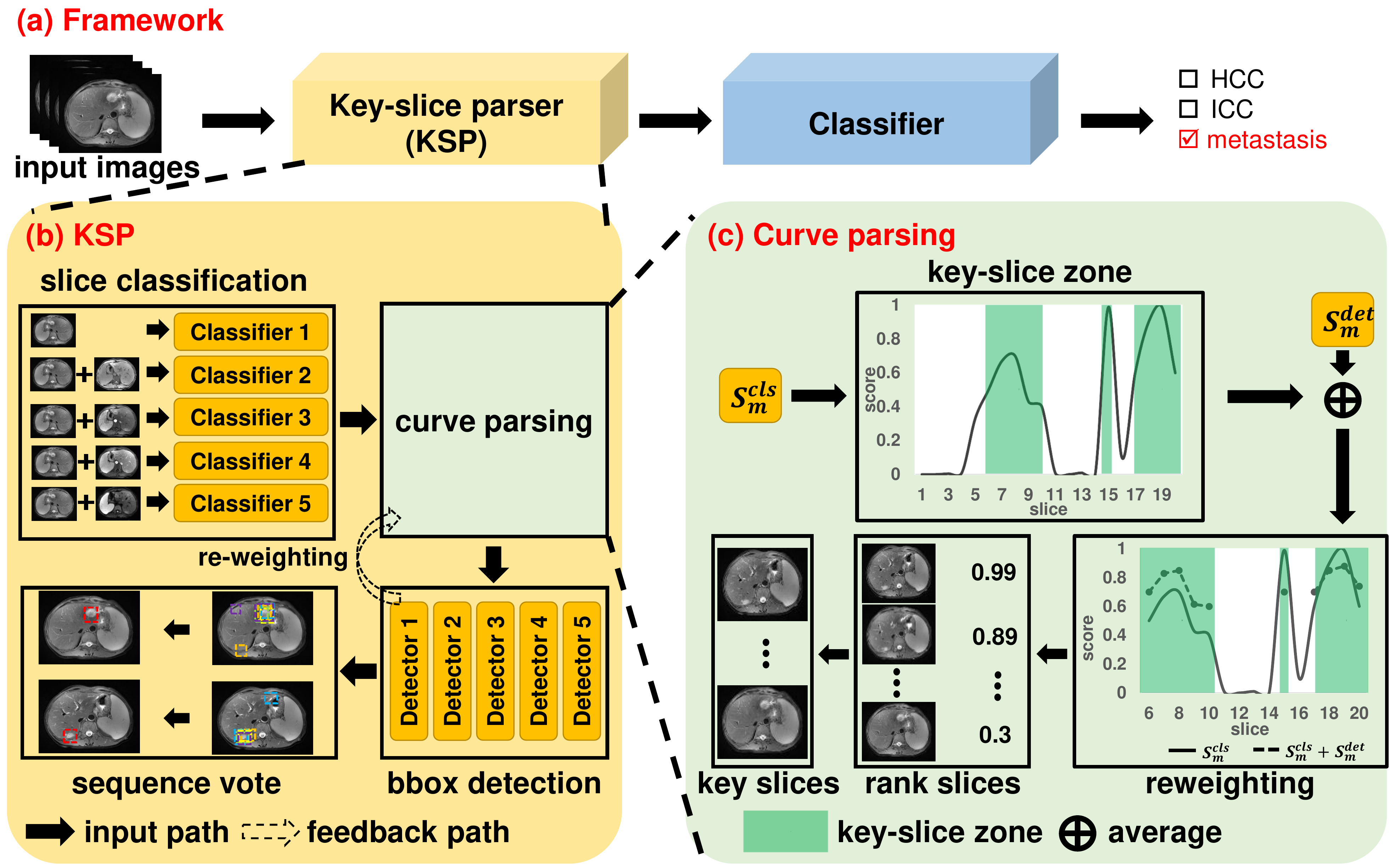}
\caption{The proposed (a) framework is composed of (b) KSP and lesion characterization. KSP consists of slice classifier, detection and (c) curve parsing.}
\label{fig: framework}
\end{figure}

\section{Methods}

\subsection{Overview}

Fig.~\ref{fig: framework}(a) illustrates our approach, which comprises a key-slice parser (KSP) and a liver lesion classifier. As illustrated and defined by Fig.~\ref{fig: tumor examples}(b), we assume we are given a dataset of MR volumes with five sequences/phases. Formally, assuming $\rm N$ studies, we define our dataset as $\mathcal{D} = \{\mathcal{X}_{i}, \, \mathcal{B}_{i}, \, y_{i}\}_{i=1}^{\rm N}$, where  $\mathcal{X}_{i}=\{X_{i,j}\}_{j=1}^5$ is the MR sequences and $y_{i}$ is a study-level lesion-type label. Lesions are either (1) annotated by 2D bounding boxes (bboxes) using RECIST-style marks~\cite{Eisenhauer2009} or (2) when they are too numerous to be individually annotated, a bbox over each cluster is provided.  See Fig.~\ref{fig: tumor examples}(a) for an illustration of the two types. Given the extreme care and multiple readers needed for lesion masks~\cite{Bilic_2019}, bbox labels are much more practical to generate. We use $m$ to represents individual slices, \eg{} $X_{i,j,m}$, which also selects any corresponding bboxes, $\mathcal{B}_{i,m}$. From the  bboxes, we can also define slices as being ``key'', ``marginal'', and ``non-key''. Marginal slices are slices within a buffer of one slice from the beginning or end of any lesion, see our supplementary for examples. We will drop the $i$ when appropriate.

\subsection{Key-Slice Parser}

Because any popular classifier can be used for lesion classification, our methodological focus is on the KSP. Illustrated in Fig.~\ref{fig: framework}(b), KSP decomposes localization the simpler problem of key-slice ranking followed by key ROI regression.

Slice ranking identifies whether each MR slice is a key slice or not. Any state-of-the-art classifier can be used, trained on ``key'' and ``non-key'' slices, with  ``marginal'' ones ignored. But care must be taken to handle multi-sequence MR data.  In short, the MR sequences or phases where lesions are visible vary, somewhat unpredictably, from lesion to lesion. EF models, \ie{} inputting a five-channel slice, can be susceptible to overfitting to the specific sequence behavior seen in the training set. Examining each MR sequence more independently mitigates this risk. Thus, we perform late fusion (LF). More specifically, because T2WI is the most informative sequence for liver tumors~\cite{aube2017easl}, we use T2WI as an anchor (T2-anchor) and pair each of the remaining sequences with it (and also one T2WI-only sequence), training a separate model for each. Unlike standard LF, the T2-anchor LF approach indeed boosts the performance over EF, see our ablation study in the supplementary. Under the T2-anchor LF approach, we obtain confidence scores for each of the five T2-anchor sequences, $j$, and for each slice, $m$: $s^{\mathrm{cls}}_{j,m}$. We average the confidence score across all sequences to compute a slice-wise classification confidence:
\begin{align}
    s^{\mathrm{cls}}_{m} =\frac{1}{5} \sum_{j=1}^5 s^{\mathrm{cls}}_{j,m}\ \mathrm{.} \label{eqn:cls_conf}
\end{align}

Our decomposition strategy means that selected key slices should contain at least one lesion. We take advantage of this prior knowledge to regress a single key ROI from each prospective key slice. To do this, we train any state-of-the-art detector on each T2-anchor sequence, producing a set of bbox confidence values and locations for each slice and sequence: $\mathcal{S}_{j,m}^{\mathrm{det}}$ and $\hat{\mathcal{B}}_{j,m}$, respectively, and we group the outputs across all sequences together: $\mathcal{S}_{m}^{\mathrm{det}},\,\hat{\mathcal{B}}_{m} = \{\mathcal{S}_{j,m}^{\mathrm{det}},\,\hat{\mathcal{B}}_{j,m}\}_{j=1}^{5}$. To produce single ROI, we use a voting scheme where for every possible pixel location we sum up the detection confidences of any bbox, $s^{\mathrm{det}}_{m,k}\in \hat{\mathcal{B}}_{m}$, that overlaps with it. We then choose the pixel location with the highest detection confidence sum. For all the bboxes that overlap the chosen pixel location, we take their mean location and size as the final slice-wise ROI. Importantly, we filter out low-confidence ROIs using a threshold $t$, which is determined by examining the overlap of resulting slice-wise ROIs with ground truth bboxes in validation. We choose the $t$ that provides the best empirical cumulative distribution function of overlaps, and thus the best balance between false negatives and false positives. 

The final step is to rank slices and  their corresponding ROIs, as illustrated in Fig.\ref{fig: framework}(c). We start by producing a confidence curve across all slices using $s^{\mathrm{cls}}_{m}$. From this curve we identify peaks, which ideally should each correspond to the presence of a true lesion. Each peak defines a key-slice zone, which is the adjoining region where confidence values are within $1/2$ of the ``peak''. Only key slices in key-slice zones will be ranked and selected, and we only admit slices that contain at least one bbox with a confidence score $>t$. 

Since good performance relies on selecting the correct slices, we use detection to build in redundancy and to better rank slices in the key-slice zone. Specifically, from all bbox confidences in a slice, $\mathcal{S}_{m}^{\mathrm{det}}$, we compute a slice-wise confidence by choosing the maximum bbox confidence. We bias these confidences toward larger bboxes, based on the assumption that they are more diagnosable:
\begin{align}
    s^{\mathrm{det}}_{m} = \max\left(\left\{(a_{m,k} + s^{\mathrm{det}}_{m,k})/2\right\}_{k=1}^{\rm{\hat{K}}_m}\right) \textrm{,} \label{eqn:det}
\end{align}
where $k$ indexes the predicted bboxes and confidences in $\mathcal{S}_{m}^{\mathrm{det}}$ and $\hat{\mathcal{B}}_{m}$ and $a_{m,k}\in (0,1]$ is the normalized bbox area across all slices.  Next, we combine classification and detection scores:
\begin{align}
    s^{\mathrm{cls+det}}_{m}=(s^{\mathrm{cls}}_{m}+s^{\mathrm{det}}_{m})/2 \mathrm{.} \label{eqn:cls+det}
\end{align}
We rank slices using \eqref{eqn:cls+det} and select the top $T\%$ of slices. We choose $T$ by examining the distribution of \emph{within-study} precision and recalls across all validation studies. We choose the $T$ giving an \emph{across-study} average recall $>=0.5$ and a first quartile (Q1) precision of $>=0.6$ (see Results). This strategy is applied for all evaluated detectors. The corresponding slice-wise ROIs comprise the key ROIs.

\section{Results}
\label{sec: Experiments}

\subsubsection{Setup}

We collected $430$ multi-phase multi-sequence MR studies ($2150$ volumes) from \emph{Anonymized} Hospital. The selection criteria was any patient who had surgical reaction or biopsy in the period between 2006 and 2019 where T1WI, T2WI, T1WI-V, T1WI-A, and DWI sequences are available. Lesion distribution was $207$, $113$, and $110$ patients with HCC, ICC, and metastasis, respectively. The data was then split patient-wise using five-fold cross validation, with $70\%$, $10\%$, and $20\%$ used for training, validation, and testing, respectively. Data splitting was executed on HCC, ICC and metastasis independently to avoid imbalanced distributions. RECIST marks were labeled on each slice, under the supervision of a hepatic physician with $>10$ years experience. From there, a bbox was generated. For clusters of lesions too numerous to individually mark, a bbox over each cluster was drawn, as shown in Fig.~\ref{fig: tumor examples}(a).  As key-slice classifier we use DenseNet121~\cite{huang2017densely}. As detectors we evaluated three KSP options: CenterNet has been used to achieve state-of-the-art results in DeepLesion~\cite{cai2020lesion};  ATSS achieves the best performance on COCO~\cite{zhang2020bridging}; and 3DCE is a powerful detector specifically designed for lesion localization~\cite{yan20183d}. All detectors were trained using the T2-anchor LF approach. Standard preprocessing and hyper-parameter settings were used for all modules, which are outlined in the supplementary.

\subsubsection{Key-Slice Selection}

\begin{figure}[t]
\centering
\includegraphics[width=1.0\columnwidth]{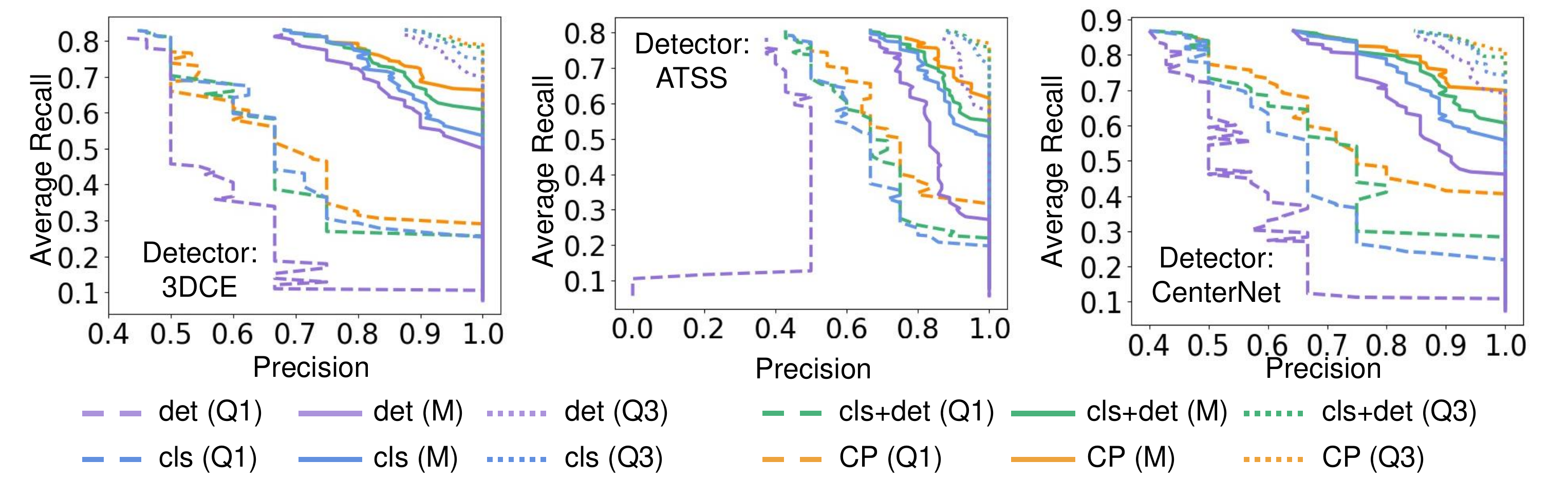}
\caption{PR curves of three key slice selection strategies. For each choice of top $T\%$ of slices selected for each study, we graph the corresponding average recall (across all patients) and the first quartile (Q1), median (M), and third quartile (Q3) precisions.}
\label{fig: slice_selection}
\end{figure}

We measure the impact of detection-based reweighting and curve parsing (CP). To do this, we rank key slices based on a) directly using detection output, \ie{} $s^{\rm{det}}_{m}$, b) directly using classification output, \ie{} $s^{\rm{cls}}_{m}$, c) using classification and detection confidences, \ie{} Eq.\eqref{eqn:cls+det}, and d) including CP in key-slice selection. As metrics, we select the top $T\%$ of ranked slices across all studies. For each choice of $T$, we calculate the within-study precision and recall, giving us a distribution of precision and recalls across studies. Thus, we graph the corresponding average recall (across patients) along with the median, first quartile (Q1) and third quartile (Q3) precision, providing typical, lower-, and upper-bound performances. In Fig.~\ref{fig: slice_selection}, detection-based re-weighting significantly outperforms detection and classification, boosting the Q1 and median precision, respectively. CP provides additional boosts in precision and recall, with notable boosts in lower-bound performance (robustness). To choose $T$ we select the value corresponding to an average recall $>=0.5$ and a Q1 precision $>=0.6$, which balances between finding all slices while keeping good precision. For CenterNet, ATSS, and 3DCE this corresponds to keeping $48\%$, $54\%$, and $50\%$ of the top slices, respectively.

\subsubsection{Localization}

\begin{figure}[t]
\centering
\includegraphics[width=0.99\columnwidth]{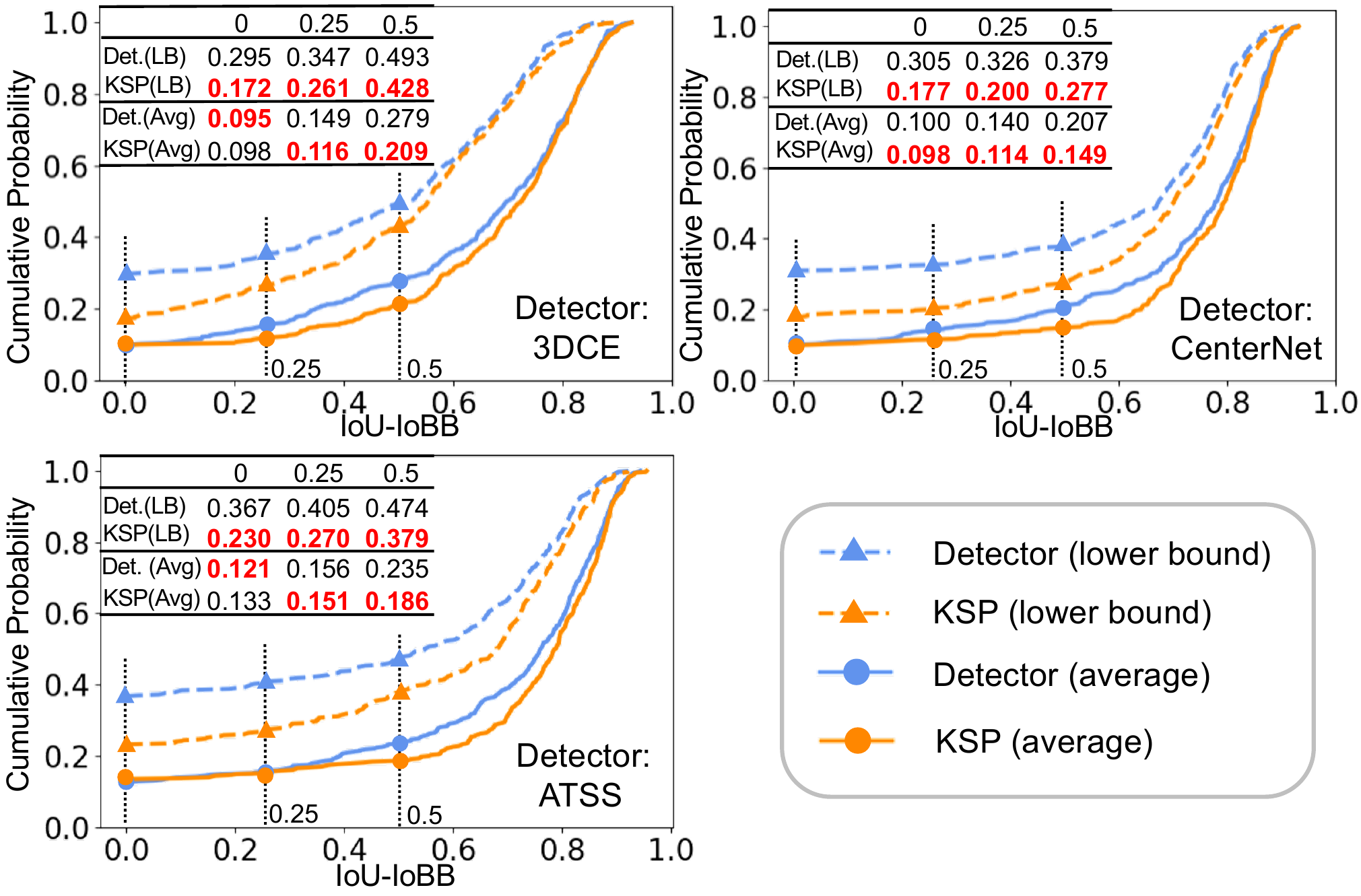}
\caption{Empirical CDF curves of three detectors and their KSP counterparts on the test set. Solid curves represent the average ROI overlap across selected slices for each patient, while dashed curves show the lower-bound (LB) overlap. Tables show exact percentages of patients with an average or LB overlap $=0$, $<=0.25$, and $<=0.50$, where lower percentages indicate better performance.}
\label{fig: cdf_test}
\end{figure}

Unlike standard detection setups, for CAD we are not interested in free response operating characteristics with arbitrary overlap cutoffs. Instead, we are only interested in whether we can select high-quality ROIs. Thus, we measure the overlap of selected ROIs against any ground truth bbox using the intersection over union (IoU). When ground truth bboxes are drawn over lesion clusters, we use the intersection over bounding box (IoBB)~\cite{wang2017chestx}  as an IoU proxy. For each patient, we examine the average overlap across all selected ROIs and also the worst case, \ie{} lower bound (LB) overlap. We then directly observe the empirical cumulative distribution function (CDF) of these overlaps \emph{across all patients}. We evaluate whether the KSP can enhance the performance of the three tested detectors, which would otherwise directly output key ROIs according to their bbox confidence scores. 

From Fig.~\ref{fig: cdf_test}, the improvements provided by KSP is apparent on all detectors. When examining the mean overlap for each patient, the percentage of patients with low overlap ($<=25\%$ IoU-IoBB) is decreased by $0.5\%\sim3.3\%$. \emph{Much more significantly}, the LB performance indicates that KSP results in roughly $13\%$ fewer patients with zero overlap and $12\%$ fewer patients with low overlap. Thus, the KSP better ensures that no poor ROIs get selected and passed on to classification. It should be noted that the LB metrics directly measure robustness, which is the main motivating reason for the KSP framework. Hence, the corresponding LB improvements validate the KSP approach of hierarchically decomposing the problem into key-slice classification and ROI regression. 

\begin{table}[t]
\caption{Lesion characterization performance. Radiomics is implemented based on the manual localization and SaDT\cite{huo2020harvesting} cannot be used without ROIs. Results when using DenseNet121 with ground truth bboxes are reported as the upper bound. The bold numbers indicate the best performance under each metric except the upper bound.}\label{tab:cls_test}
\setlength{\tabcolsep}{0.3mm}
\begin{tabular}{l|c|c|c|c|c}
\toprule
Methods & Accuracy & mean F1  & F1(HCC)  & F1(ICC)  & F1(Meta.)\\
\midrule
Radiomics\cite{van2017computational}       & $58.65\pm4.51$ & $55.07\pm6.34$ & $71.37\pm4.05$ & $39.50\pm10.94$ & $54.36\pm9.26$ \\
\midrule
ResNet101\cite{he2016deep}       & $59.54\pm3.53$ & $50.03\pm3.55$ & $76.85\pm4.55$ & $12.74\pm8.58$ & $60.51\pm4.55$ \\
DenseNet121\cite{huang2017densely}     & $61.68\pm5.43$ & $51.24\pm3.73$ & $75.91\pm10.37$ & $17.54\pm14.51$ & $50.15\pm7.57$ \\
ResNeXt101\cite{xie2017aggregated}      & $60.51\pm5.32$ & $54.96\pm5.60$ & $78.15\pm6.01$ & $27.62\pm11.89$ & $59.12\pm6.79$ \\
DeepTEN\cite{zhang2017deep}         & $53.97\pm3.38$ & $54.74\pm3.12$ & $65.03\pm6.60$ & $41.39\pm5.63$ & $57.80\pm6.40$ \\
\midrule
KSP+ResNet101   & $64.91\pm6.42$ & $61.32\pm5.91$ & $77.00\pm6.09$ & $45.73\pm6.72$ & $61.26\pm6.44$ \\
KSP+DenseNet121 & $\mathbf{69.62\pm3.13}$ & $\mathbf{66.49\pm2.78}$ & $80.12\pm3.54$ & $\mathbf{55.34\pm4.88}$ & $\mathbf{64.02\pm6.81}$ \\
KSP+ResNeXt101  & $67.26\pm4.18$ & $62.82\pm3.98$ & $\mathbf{80.33\pm3.54}$ & $45.86\pm4.15$ & $62.27\pm7.98$ \\
KSP+DeepTEN     & $67.20\pm2.79$ & $64.14\pm3.95$ & $77.07\pm3.26$ & $51.61\pm11.88$ & $63.74\pm8.48$ \\
KSP+SaDT        & $67.26\pm3.91$ & 63.25$\pm3.63$ & $79.69\pm3.62$ & $49.26\pm8.51$ & $60.80\pm8.11$ \\
\midrule
Upper bound     & $70.68\pm2.97$ & $68.02\pm3.83$ & $78.68\pm3.61$ & $57.59\pm7.41$ & $67.79\pm7.78$\\
\bottomrule

\end{tabular}
\end{table}

\subsubsection{Characterization}
 Finally, for the overall lesion characterization performance, we measure patient-wise accuracy, one-vs-all and mean F1 score(s) of the three tumor types, with emphasis on HCC-vs-others given its prominence in clinical work~\cite{aube2017easl}. According to Fig.~\ref{fig: cdf_test}, CenterNet with KSP surpasses 3DCE and ATSS in average and LB overlap, respectively. Therefore, we choose it as our KSP detector and train and test various classifiers on its ROIs.  Patient-wise diagnoses are produced by averaging classifications from detected ROIs weighted by confidence. As demonstrated in Table~\ref{tab:cls_test}, compared with using classifiers alone, KSP significantly improves accuracy (+5\%$\sim$+8\%), mean F1 (+8\%$\sim$+15\%) and HCC F1 scores (+0.15\%$\sim$+12\%). DenseNet121~\cite{xie2017aggregated} performs best, garnering  an HCC vs. others F1 score of 0.801, which is comparable to reported physician performance (0.791)~\cite{aube2017easl}. In addition, we also produced an upper bound by testing DenseNet121 on oracle tumor locations, and there is only a marginal gap between it and our best results---$1\%$ in accuracy and $1.5\%$ in mean F1 score. This further validates the effectiveness of KSP, suggesting that performance bottlenecks may now be due to classifier limitations, which we leave for future work.

\section{Discussion and Conclusion}
\label{sec: Discussion}

As the medical image field progresses toward more clinically viable CAD, research efforts will likely increasingly focus on ensuring true robustness. We contribute toward this goal for liver lesion characterization. Specifically, we articulate a physician-inspired decompositional approach toward ROI localization that breaks down the complex problem into key-slice identification and then ROI regression. Using our proposed framework, our KSP realization can achieve very high robustness, with $87\%$ of its ROIs having an overlap of $>=40\%$. Overall, our fully automated CAD solution can achieve an HCC-vs-others F1 score of $80.1\%$. Importantly, this performance is reported on histopathologically-confirmed cases, which selects for the most challenging cases requiring CAD intervention. Even so, this matches reported clinical performances of $79.1\%$~\cite{aube2017easl}, despite such studies including both radiologically and histopathologically-confirmed cases. Given the challenging nature of liver lesion characterization, our proposed CAD system represents a step forward toward more clinically practical solutions. 
%
%
%
\bibliographystyle{splncs04}
\bibliography{mybibliography}

\newpage
~\\
\centerline{\Large \textbf{Supplementary Material}}

\setcounter{section}{0}  

\section{Definition of Key Slices}
\begin{figure}[htbp]
\centering
\includegraphics[width=0.99\columnwidth]{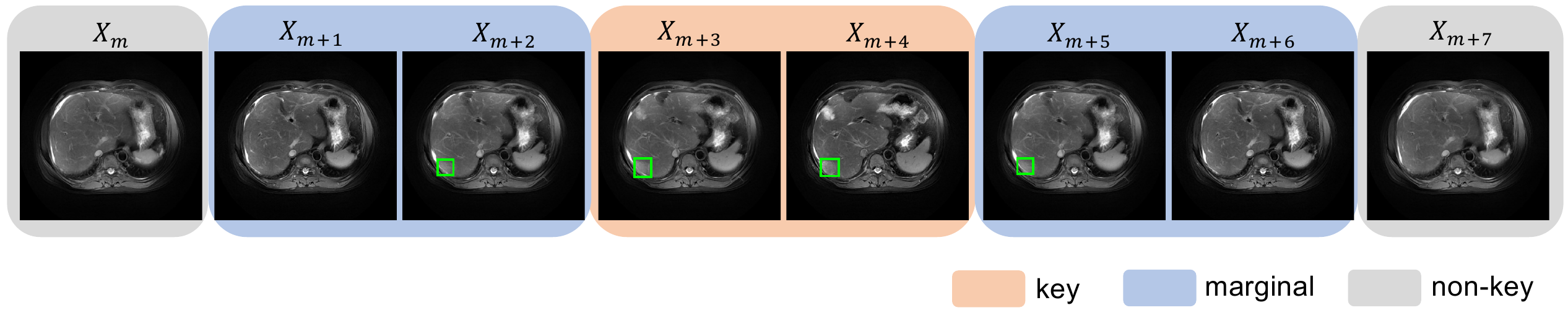}
\caption{The definition of key, marginal and non-key slices. Eight consecutive slices are demonstrated with ground truth bboxes showing the location of tumors.}
\label{fig: keyslice}
\end{figure}
We label all slices as ``key'', ``marginal'', and ``non-key'' using the scheme illustrated in Fig.~\ref{fig: keyslice}. For cases with a single tumor in the liver, ``marginal'' slices are those within a buffer of one slice from the beginning or end of a lesion. Then slices with tumors between two ``marginal'' regions are defined as ``key'' slices and the remaining slices are defined as ``non-key'' slices. For cases with more than one tumor, we merge the designations of each individual lesion together to create a slice-wise label. Under this protocol, a ``key'' slice is any slice that captures one or more individual lesion ``key'' slices. In the remaining slices, a ``marginal'' slice is any slice that captures one or more individual lesion ``marginal'' slices. Finally, for those that are not defined as ``key'' or ``marginal'', they are treated as ``non-key'' slices.

\section{Implementation Details}

\textbf{Preprocessing} We resampled all MR volumes and aligned them using the DEEDS algorithm~\cite{heinrich2013mrf}. All volumes were preprocessed by clipping within the $0.1\%$ and $99.9\%$ percentile values. For all experiments we augmented the data by random  rotations and gamma intensity transforms.  

\textbf{Slice classification in KSP} used a DenseNet121~\cite{huang2017densely} backbone. Following \cite{yan2018unsupervised}, we add an additional $1\times1$ convolutional layer before global pooling and use log-sum-exp (LSE) pooling, finding it outperforms the standard average pooling. Three adjacent slices are inputted to provide some 3D context. The batch size is set as $20$, an Adam optimizer \cite{kingma2014adam} with initial learning rate as $1\times10^{-4}$ is used. The learning rate is decayed by $0.01$ after every 1000 iterations. 

For the \textbf{detection in KSP}, we demonstrate the benefits of our framework on three advanced detectors - 3DCE~\cite{yan20183d}, CenterNet~\cite{zhou2019objects} and ATSS~\cite{zhang2020bridging}. To avoid overly tuning hyper-parameters, the network structure and loss use the same settings as the original papers. The batch size and learning rate are set as 30 and $1\times10^{-4}$, respectively, which follows linear learning rate rule~\cite{Goyal2017AccurateLM}. Each model is trained for 50 epochs. Random scaling and cropping was added as data augmentation for all tested detectors. 

As for \textbf{lesion characterization}, we test radiomics~\cite{van2017computational}, three standard classifiers, ResNet101~\cite{he2016deep}, DenseNet121~\cite{huang2017densely} and ResNeXt101~\cite{xie2017aggregated}, as well as two texture based classifiers, DeepTEN~\cite{zhang2017deep} and SaDT~\cite{huo2020harvesting}. In radiomics, the support vector machine (SVM) classifier is implemented with extracted features from manually localized tumors, including shape, first order statistics, neighboring gray level dependence method (NGLDM)~\cite{sun1983neighboring}, gray level size zone matrix (GLSZM)~\cite{thibault2013shape}, gray level run length matrix (GLRLM)~\cite{galloway1975texture} and gray-level co-occurrence matrix (GLCM)~\cite{haralick1973textural}. As for deep-learning-based classifiers, the batch size and learning rate are set as 8 and $1\times10^{-4}$, respectively. Networks are trained based on ground truth bounding boxes (bboxes) for 70 epochs. Random rotation, scaling and cropping are adopted as augmentation. Besides, ground truth bboxes are randomly shifted and resized to simulate the imperfect localization.

\section{Evaluation of Different Fusion Methods}
\begin{figure}[tbp]
\centering
\includegraphics[width=0.99\columnwidth]{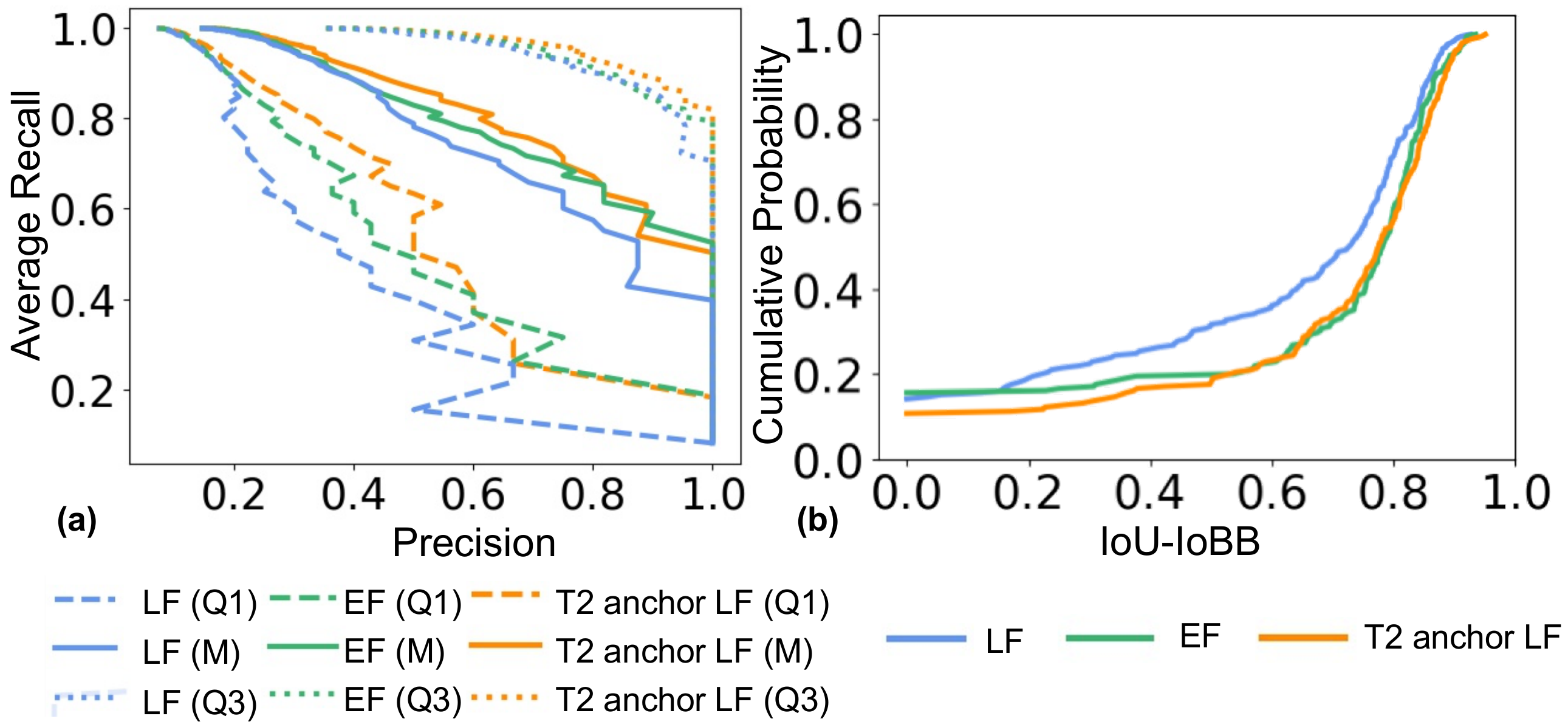}
\caption{(a) The comparison of early fusion (EF), standard late fusion (LF), and T2-anchor LF on key-slice classification. Average recall (across all patients) and the first quartile (Q1), median (M), and third quartile (Q3) precisions are graphed. (b) The CDF curves of three fusion approach on detection.}
\label{fig: fusion}
\end{figure}

In Fig.~\ref{fig: fusion}(a), we measure three fusion approaches, including standard late fusion (LF), early fusion (EF), and also T2-anchor LF on key-slice classification. As metrics shows, when keeping the same average recall, T2-anchor LF outperforms both LF and EF in the first quartile (Q1), medium (M) and third quartile (Q3) precision, showing its superiority of localizing key slices. In Fig.~\ref{fig: fusion}(b), the cumulative probability curves of CenterNet~\cite{zhou2019objects} with three fusion methods are demonstrated. To evaluate the detection performance independently, the curves show results directly from the detector without key-slice classification. The curve of T2-anchor LF is lower than the other two especially when IoU-IoBB is smaller than 0.5. This indicates T2-anchor LF surpasses LF and EF by decreasing the ratio of bboxes of low overlaps with ground truth.

\end{document}